\documentclass{bmvc2k}

\usepackage{graphicx}
\usepackage{amssymb}
\usepackage{amsmath,amsfonts,amsthm,bm} 
\usepackage{hyperref}
\usepackage{url}
\usepackage{mathtools}
\usepackage{subfigure}
\usepackage{wrapfig}
\usepackage[ruled,vlined,linesnumbered]{algorithm2e}
\SetAlFnt{\small}
\SetAlCapFnt{\small}
\SetAlCapNameFnt{\small}
\usepackage{caption, floatrow}
\DeclareFloatVCode{myrowsep}{\vskip 4ex}
\usepackage{float}
\newsavebox{\largestimage}
\makeatletter
\newcommand{\removelatexerror}{\let\@latex@error\@gobble}
\makeatother
\everypar\expandafter{\the\everypar\looseness=-1 }

\title{Self-Paced Learning with \\ Adaptive Deep Visual Embeddings}

\addauthor{Vithursan Thangarasa$^{1,}$}{vthangar@uoguelph.ca}{2}
\addauthor{Graham W. Taylor$^{1,2,}$}{gwtaylor@uoguelph.ca}{3}

\addinstitution{
School of Engineering\\
University of Guelph\\
Guelph, Canada
}
\addinstitution{
	Vector Institute for Artificial Intelligence\\
	Canada
}
\addinstitution{
Canadian Institute for \\ Advanced Research (CIFAR)
}

\runninghead{Thangarasa, Taylor}{SPL with Adaptive Deep Visual Embeddings}

\begin{document}

\maketitle

\begin{abstract}
Selecting the most appropriate data examples to present a deep neural network
(DNN) at different stages of training is an unsolved challenge. Though
practitioners typically ignore this problem, a non-trivial data scheduling
method may result in a significant improvement in both convergence and
generalization performance. In this paper, we introduce Self-Paced Learning
with Adaptive Deep Visual Embeddings (SPL-ADVisE), a novel end-to-end training
protocol that unites self-paced learning (SPL) and deep metric learning (DML).
We leverage the Magnet Loss to train an \emph{embedding} convolutional neural
network (CNN) to learn a salient representation space. The \emph{student} CNN
classifier dynamically selects similar instance-level training examples to form
a mini-batch, where the \textit{easiness} from the cross-entropy loss and the
\textit{true diverseness} of examples from the learned metric space serve as
sample importance priors. To demonstrate the effectiveness of SPL-ADVisE, we
use deep CNN architectures for the task of supervised image classification on
several coarse- and fine-grained visual recognition datasets. Results show
that, across all datasets, the proposed method converges faster and reaches a
higher final accuracy than other SPL variants, particularly on fine-grained
classes.
\end{abstract}

\section{Introduction}\label{sec:intro}
The standard method for training deep neural networks (DNNs) is stochastic
gradient descent (SGD), which employs backpropagation to compute gradients. It
typically relies on fixed-size mini-batches of random samples drawn from a
finite dataset. However, the contribution of each sample during model training
varies across training iterations and configurations of the model's
parameters~\citep{lapedriza2013all}. This raises the importance of~\emph{data
scheduling} for training DNNs; that is, searching for an optimal ordering
of training examples that are presented to the model. Previous studies on
curriculum learning (CL)~\citep{Bengio:2009:CL:1553374.1553380} show that
organizing training samples based on the ascending order of difficulty can
be favourable for model training. However, in CL, the curriculum remains fixed over the
iterations and is determined without any knowledge or introspection of the
model's learning. Self-paced learning \citep{Kumar:2010:SLL:2997189.2997322}
presents a method for dynamically generating a curriculum by biasing samples
based on their \textit{easiness} under the current model parameters. This can
lead to a highly imbalanced selection of samples, i.e.~very few instances of
some classes are chosen, which negatively affects the training process due to
overfitting. \citet{DBLP:journals/corr/LoshchilovH15} propose a simple batch
selection strategy based on the loss values of training data for speeding up
neural network training. However, the approach is computationally expensive and
the results are inconclusive, as it achieves high performance on MNIST, but
fails on CIFAR-10. Their work reveals that selecting the examples to present to
a DNN is non-trivial, yet the strategy of uniformly sampling the training data
set is not necessarily the optimal choice.

\citet{Jiang2014SelfPacedLW} show that partitioning the data into groups with
respect to \textit{diversity} and \textit{easiness} in their self-paced
learning with diversity (SPLD) framework, can have a substantial effect on
training. Rather than constraining the model to limited groups, they propose to
spread the sample selection as wide as possible to obtain diverse samples of
similar easiness.  However, their use of $K$-Means and Spectral Clustering to
partition the data into groups can lead to sub-optimal clustering results when
learning non-linear feature representations. Therefore, learning an appropriate
metric by which to capture similarity among arbitrary groups of data is of great
practical importance. Deep metric learning (DML) approaches have recently
attracted considerable attention and have been the focus of numerous
studies~\citep{bell15productnet,Schroff2015FaceNetAU}. The most common methods
are supervised, where a feature space in which distance corresponds to class
similarity is obtained. The Magnet Loss~\citep{DBLP:journals/corr/RippelPDB15}
presents state-of-the-art performance on fine-grained classification
tasks. \citet{Song2017DeepML} show that it achieves state-of-the-art on
clustering and retrieval tasks.

This paper makes two key contributions toward scheduling data examples in the
mini-batch setting:~(1) We propose a general sample selection framework called
Self-Paced Learning with Adaptive Deep Visual Embeddings (SPL-ADVisE) that is
independent of model architecture or objective, and learns when to introduce
certain samples to the DNN during training.~(2) To our knowledge, we are the
first to leverage metric learning to improve SPL. We exploit a
new type of knowledge \textemdash similar instance-level samples are discovered
through an \textit{embedding} network trained by DML concurrently with the
self-paced learner.

\section{Related Work}

\noindent \textbf{Learning Small and Easy}.
The perspective of ``starting small and easy'' for structuring the learning
regimen of neural networks dates back decades to~\citep{Elman1993-ELMLAD}.
Recent studies show that selecting a subset of
\textit{good} samples for training a classifier can lead to better
results than using all the samples~\citep{Lee2011LearningTE,lapedriza2013all}.
Pioneering work in this direction is CL~\citep{Bengio:2009:CL:1553374.1553380},
which introduced a heuristic measure of easiness to determine the selection of
samples from the training data. By comparison,
SPL~\citep{Kumar:2010:SLL:2997189.2997322} quantifies the easiness by the
current sample loss. The training instances with loss values larger than a
threshold, $\lambda$, are neglected during training and $\lambda$
dynamically increases in the training process to include more complex samples,
until all training instances are considered. Previous studies prove that CL and
SPL strategies are instrumental in avoiding bad local minima and improving
generalization~\citep{Khan2011HowDH,Glehre2016KnowledgeMI,DBLP:conf/atal/Peng17}.
This theory has been widely applied to various problems, including visual
tracking~\citep{huang2017self}, medical imaging analysis~\citep{li2017self},
and person re-identification~\citep{Zhou2017DeepSL}. In CL, the increasing
entropy theory shows that a curriculum should progressively increase the
diversity of training examples during learning. An effective SPL strategy should generate a curriculum that includes easy and diverse
examples that are sufficiently dissimilar from what a DNN has already learned.
In SPLD~\citep{Jiang2014SelfPacedLW}, training data are pre-clustered in order
to balance the selection of the easiest samples with a sufficient inter-cluster
diversity. However, the clusters and the feature space are fixed: they do not
depend on the current self-paced training iteration. Adaptation of this method
to a deep learning scenario, where the feature space changes during learning,
is non-trivial. Our self-paced sample selection framework targets a similar
goal but obtains diversity of samples through a DML approach to adaptively
sculpt a representation space by autonomously identifying and respecting
intra-class variation and inter-class similarity.\\

\noindent \textbf{Learning the Metric Space}.
Deep metric learning has gained much popularity in recent years, along with the
success of deep learning. The objective of DML is to learn a distance metric
consistent with a given set of constraints, which usually aim to minimize the
distances between pairs of data points from the same class and maximize the
distances between pairs of data points from different classes. DML approaches
have shown promising results on various tasks, such as zero-shot
learning~\citep{41869}, face recognition~\citep{Schroff2015FaceNetAU}, and
feature matching~\citep{choy_nips16}. DML can also be used for challenging,
extreme classification settings, where the number of classes is very large and
the number of examples per class becomes scarce. Most DML methods
define the loss in terms of pairs~\citep{songCVPR16},
triplets~\citep{Wang2017DeepML} or $n$-pair tuples~\citep{Sohn2016ImprovedDM}
inside the training mini-batch. These methods require a separate data
preparation stage, which has a very expensive time and space cost. Also, they do
not take the global structure of the embedding space into consideration, which
can result in reduced clustering. An alternative is the Magnet
Loss~\citep{DBLP:journals/corr/RippelPDB15}, DML via Facility
Location~\citep{Song2017DeepML}, and Class-Conditional Metric Learning
(CCML)~\citep{Im2016ccml}, which do not require the training data to be
preprocessed in rigid paired format and are aware of the global structure
of the embedding space. Our work employs the Magnet Loss to learn a
representation space, where we compute centroids on the raw features and then
update the learned representation continuously.\looseness=-1

To our knowledge, the concept of employing DML for SPL-based DNN training has
not been investigated. Effectively, a deep CNN trained by the Magnet Loss
learns a representation in which a clustering of that space represents the
diversity of the data. A student model exploits this space by forming dynamic
mini-batches that select samples based on \textit{true diverseness} defined by
that clustering and \textit{easiness} defined by the student's current loss.
Our architecture combines the strength of adaptive sampling, the efficiency of
mini-batch online learning, and the flexibility of representation learning to
form an effective self-paced strategy in an end-to-end DNN training protocol.

\section{Proposed Method}
\begin{figure*}[t]
\captionsetup{font=small}
\begin{center}
\includegraphics[width=1.0\linewidth]{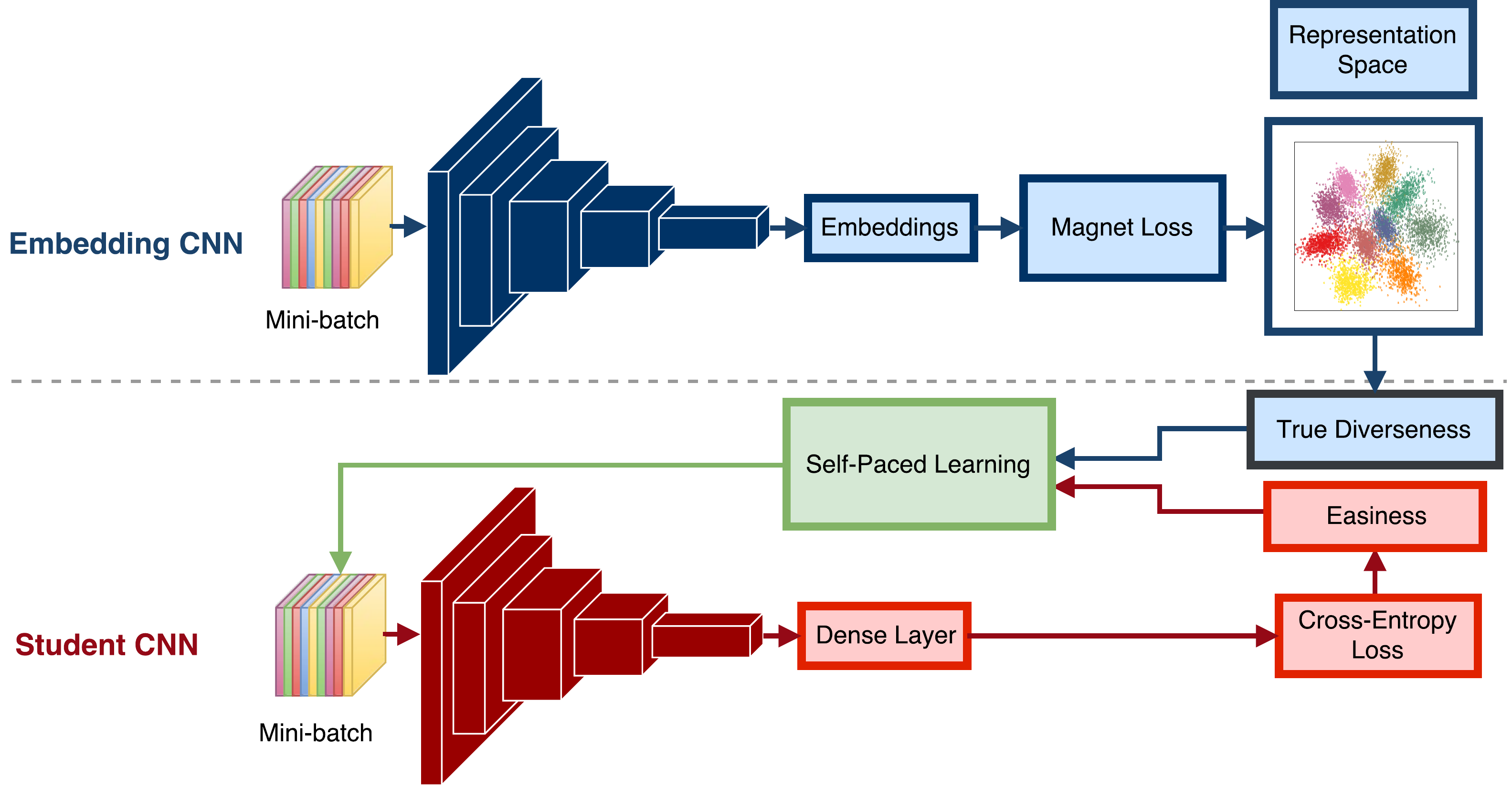}
\end{center}
\caption{The SPL-ADVisE framework, consisting of an embedding CNN that learns a
representation for the student CNN to create a self-paced strategy based on
\textit{easiness} and \textit{true diverseness} as sample importance priors.}
\label{fig:arch}
\end{figure*}

The Self-Paced Learning with Adaptive Deep Visual Embeddings (SPL-ADVisE)
framework consists of dual complementary DNNs. An \textit{embedding} DNN learns
a salient representation space, then transfers its knowledge to the self-paced
selection strategy for training the second DNN, called the \emph{student}. In
this work, but without loss of generality, we focus on training deep CNNs for the task of supervised image
classification. More specifically, an \textit{embedding} CNN is trained
alongside the \textit{student} CNN of ultimate interest
(Figure~\ref{fig:arch}). In this framework, we want to form mini-batches
using the \textit{easiness} and \textit{true diverseness} as sample importance
priors for the selection of training samples. Given that we are learning the
representation space adaptively alongside the student as training progresses,
this has negligible computational cost compared to the actual training of the
student CNN (see Section~\ref{sec:experiments}).\\

\subsection{Embedding CNN}
Though there are many possible DML frameworks from which to choose, we adopt
the Magnet Loss~\citep{DBLP:journals/corr/RippelPDB15} because of its strong
empirical gains relative to popular baselines such as the margin-based Triplet
loss and softmax regression. Assuming we have a training set with $N$
input-label pairs $\mathcal{D} = \{\mathbf{x}_n, y_n\}_{n=1}^{N}$, the Magnet Loss
learns the distribution of distances for each example, to $K$ clusters
assigned for each class, $c$, denoted as $\{\mathcal{I}_{k}^{c}\}_{k=1}^{K} \
\forall \ C$ classes. The mapping of inputs to representation space are
parameterized by $\textbf{f}(\cdot;\Theta)$, where their representations are
defined as $\textbf{r}_n = \{\textbf{f}(x_n;\Theta)\}_{n=1}^{N}$. The cluster
assignments are then repositioned using an intermediate
$K$-Means++ clustering~\citep{Arthur:2007:KAC:1283383.1283494}. Therefore, for
each class $c$, we have,
\begin{align}
\{\mathcal{I}_{k}^{c}\}_{k=1}^{K} &=
\mathop{\text{argmin}}\limits_{I_1^c,\ldots,I_K^c}\sum_{k=1}^{K}\sum_{r \in
I_k^c}\|\textbf{r}-\mu_k^c\|_2^2, \nonumber \\
\mu_k^c &= \frac{1}{\|I_k^c\|}\sum_{\textbf{r} \in I_k^c} \textbf{r}. \nonumber
\end{align}
The class and assigned cluster centre of  $\textbf{r}$ are defined as
$C(\textbf{r})$ and $\mu_k^c$, respectively. The optimized indices are
represented by $\mathcal{I}$, while the indices that are the variates of the
optimization are represented by $I$. In the training procedure, the embedding
CNN constructs mini-batches with neighbourhood sampling, where we sample a local
neighbourhood at each iteration.

First, a seed cluster is sampled~$I_1 \sim
p_{\mathcal{I}}(\cdot)$, and then the nearest $M-1$ imposter
clusters~$\{I_{m}\}_{m=2}^{M}$ are fetched for $I_1$, where $M$ number of
mini-batch clusters is chosen to be $M \leq K$. The objective of learning the
metric is to ensure that nearest neighbours that belong to the same class
are clustered together, while imposters are moved away by a large margin.
Imposter clusters are considered to be groups of points that have nearest
neighbours with different labels,~i.e. clusters of examples from different
classes. Finally, $B$ training
instances~$\{x_{b}^m\}_{b=1}^B \sim p_{I_m}(\cdot)$ are uniformly sampled for
each cluster $\{I_m\}_{m=1}^{M}$. The $p_{\mathcal{I}}(\cdot)$ and
$p_{I_m}(\cdot)$ choices enable us to adapt to the current distributions of each
example in the representation space. Therefore, during training, contested
neighbourhoods with large cluster overlap can be targeted and reprehended.

The losses of each example are stored and the
average loss $\mathcal{L}$ is computed during training.
Hence, the stochastic approximation of the Magnet Loss can be more formally described as:
\begin{align}
  \label{eq:magnet_loss}
  \mathcal{L}(\Theta) = \frac{1}{MB}\sum_{m=1}^{M}\sum_{b=1}^{B}
  \Bigg\{-\log \bigg(\frac{e^{-\frac{1}{2\sigma^2}\|\textbf{r}_b^m -
  \mu_m\|_2^2-\alpha}}{\sum_{\mu:C(\mu)\neq
  C(\textbf{r}_b)}e^{-\frac{1}{2\sigma^2}\|\textbf{r}_b^m - \mu
  \|_{2}^{2}}}\bigg)
  \Bigg\}_{+},
\end{align}
where $\{\cdot\}_{+}$ is the hinge function, $\alpha \in \mathbb{R}$ is the
desired separation parameter, the cluster mean is~$\mu_m =
\frac{1}{B}\sum_{b=1}^{B}\textbf{r}_b^m$, and the variance of all samples from
their respective centers is given by $\sigma =
\frac{1}{MB-1}\sum_{m-1}^{M}\sum_{b=1}^{B}\|\textbf{r}_b^m - \mu_m\|_2^2$. A
\textit{K}-Means cluster index is used to capture the distribution of $c$ in the
embedding space for all classes $C$. A cluster refresh interval $R$ is used to
reinitialize each index using \textit{K}-Means++. We pause the embedding CNN
training and perform forward passes of all inputs to get the representations
$\textbf{r}_n$ to be indexed. The cost of \textit{K}-Means++ clustering is
negligible because the cluster index is not refreshed frequently and we only
run forward passes of the inputs~\citep{DBLP:journals/corr/RippelPDB15}. The
process for training the embedding CNN is provided in
Algorithm~\ref{alg:embeddingcnn_magnet}.

\begin{figure}[H]
  \removelatexerror
  \begin{algorithm}[H]
  \DontPrintSemicolon
  \SetAlgoLined
  \SetKwInOut{Input}{Input}\SetKwInOut{Output}{Output}
  \Input{$\mathcal{D}$ (training data), $B$ (per-cluster batch size), $M$ (mini-batch clusters), \\ $C$ (classes), $E$ (embedding CNN iterations), $R$ (cluster refresh interval)}
  \Output{A representation space $\mathcal{D}_1^1,\ldots,\mathcal{D}_K^C$.}
  \BlankLine

  Extract initial representations $\textbf{r}_n = \{\textbf{f}(x_n;\Theta)\}_{n=1}^N$.\\
  Initialize cluster assignments $\{\mathcal{I}_{k}^{c}\}_{k=1}^{K} \ \forall \  C$ using \textit{K}-Means++. \\
  \For{each iteration $\{e = 1,\ldots,E\}$}
  {
    Sample seed cluster $I_1 \sim p_{\mathcal{I}}(\cdot)$. \\
    Fetch impostor clusters $\{I_m\}_{m=2}^{M}$ for $I_1$. \\
    Uniformly sample $\{x_b^m\}_{b=1}^{B}\sim p_{I_m}(\cdot)$ for each $m$, to form mini-batch.\\
    Extract mini-batch representations $\textbf{r}_b^m = \{\textbf{f}(x_b^m;\Theta)\}_{b=1}^{B}$ for each $m$. \\
    Compute $\mathcal{L}(\Theta)$ as in Eq. \ref{eq:magnet_loss}. \\
    \If{e == R}{
        Repeat~\textbf{steps $1-2$}. \\
      }
  }
  \Return $\mathcal{D}_1^1,\ldots,\mathcal{D}_K^C$
  \caption{Train Embedding CNN with Magnet Loss.}
  \label{alg:embeddingcnn_magnet}
  \end{algorithm}
\end{figure}

\subsection{SPL-ADVisE Framework}
The aim of SPL-ADVisE can be formally described as follows. Let us assume that
a training set $\mathcal{D}$ consisting of $N$ examples, $\mathcal{D} =
\{\mathbf{x}_n\}_{n=1}^{N}$ is grouped into $K$ clusters for $C$ classes using
Algorithm~\ref{alg:embeddingcnn_magnet}. Therefore, we have
$\{\mathcal{D}^{k}\}_{k=1}^K$, where $\mathcal{D}^{k}$ corresponds to the
$k^{th}$ cluster, $n_k$ is the corresponding number of examples in that cluster and
$\sum_{k=1}^{K}n_k = N$.
The weight vector is denoted accordingly as $\{\mathcal{W}^{k}\}_{k=1}^{K}$,
where $\mathcal{W}^{k} = (\mathcal{W}_1^k,\ldots,\mathcal{W}_{n_k}^k)^T \in
[0,1]^{n_k}$. Non-zero weights of $\mathcal{W}$ are assigned to samples that
the student model considers ``easy'' and non-zero elements are distributed
across more clusters to increase diversity. Thus, in the SPL-ADVisE framework, we optimize the following objective:
\begin{align}
\min_{\theta, \mathcal{W}} \mathbb{E}(\theta,
\mathcal{W}; \lambda, \gamma) &= \sum_{i=1}^{N}\mathcal{W}_i\mathcal{L}_{ce}(y_i,
f(x_i,\theta)) -\lambda \sum_{i=1}^{N}\mathcal{W}_i - \gamma\|\mathcal{W}\|_{2,1}, \ \text{s.t} \
\mathcal{W} \in [0,1]^{N},\label{eqn:spladvise}
\end{align}
where $\mathcal{L}_{ce}$ is the cross-entropy loss, and $\lambda$, $\gamma$ are
the two pacing parameters for sampling based on
\textit{easiness} and the \textit{true diverseness} as sample importance
priors, respectively. The negative $l_1$-norm: $-\|\mathcal{W}\|_1$ is used to
select easy samples over hard samples, as seen in conventional SPL. The
negative $l_{2,1}$-norm inherited from the original SPLD algorithm is used to
disperse non-zero elements of $\mathcal{W}$ across a large number of clusters
to obtain a diverse selection of training samples. Therefore, the diversity term
$-\| \mathcal{W} \|_{2,1}$ is defined as $- \sum_{k=1}^{K}\| \mathcal{W}^{k}\|_2$.
The student CNN receives the up-to-date model parameters $\theta$, a
diverse cluster of samples, $\lambda$, $\gamma$ and outputs the optimal  $\mathcal{W}$ of
$\displaystyle \min_{\mathcal{W}}\mathbb{E}(\theta, \mathcal{W};
\lambda, \gamma)$ for extracting the global optimum of this optimization
problem. The detailed algorithm to train the student CNN with SPL-ADVisE is
presented in Algorithm \ref{alg:studentcnn_leap}.

\begin{figure}[H]
  \removelatexerror
  \begin{algorithm}[H]
  \DontPrintSemicolon
  \SetAlgoLined
  \SetKwInOut{Input}{Input}\SetKwInOut{Output}{Output}
  \Input{$\mathcal{D}$ (training data), $E^\prime$ (student CNN iterations), $\{\beta_1,  \beta_2\}$ (pace adaptation parameters)}  \Output{The model parameters $\theta$.}
  \BlankLine
  Initialize $\mathcal{W}^{*}, \lambda, \gamma$ ; \\

  Update embedding CNN using \textbf{Algorithm~\ref{alg:embeddingcnn_magnet}}, in parallel with \textbf{steps} 3-16. \\
  \For{each iteration $\{{e^\prime = 1,\ldots,E^\prime}\}$}
  {

    $\theta^{*} = \mathop{\text{argmin}}_{\theta}\mathbb{E}(\theta, \mathcal{W}^{*};\lambda, \gamma)$. \CommentSty{// Train Student CNN model.} \\
      \CommentSty{// Solve $\min_{\mathcal{W}}\mathbb{E}(\theta, \mathcal{W};
      \lambda, \gamma)$} \\
      Get clusters $\{D^{k}\}_{k=1}^{K}$ from the embedding CNN. \\
      \For{each cluster $\{k = 1,\ldots,K\}$}
      {
        Sort training samples in increasing order of $\mathcal{L}$.\\
        Represent the labels of $D^{k} \ \text{as} \ \{y_1^k,\ldots,y_{n_k}^k\}$  and weights as $\{\mathcal{W}_1^k,\ldots,\mathcal{W}_{n_k}^k\}$. \\
        \For{each sample $i = \{1,\ldots,n_{k}\}$}
        {
          \lIf{$\mathcal{L}_{ce}(y_i^{k}, f(x_i^{k},\theta^*)) < (\lambda + \gamma\frac{1}{\sqrt{i}+\sqrt{i-1}})$}
          {
            $\mathcal{W}_i^{k} = 1$ ;
          }
          \lElse
          {
            $\mathcal{W}_i^{k} = 0$ ;
          }
        }
    }
    $\mathcal{W}^{*}$ = $\mathcal{W}$; $\ $ $\theta = \theta^{*}$; \\
    $\lambda \leftarrow \beta_1\lambda$; $\ $ $\gamma \leftarrow \beta_2\lambda$; \CommentSty{// Update pace.}\\
  }
  \caption{Train Student CNN with SPL-ADVisE.}
  \label{alg:studentcnn_leap}
  \end{algorithm}
\end{figure}
\vspace{5mm}
For any $\theta$, we show how we can solve for the global optimum to
$\displaystyle \min_{\mathcal{W}}\mathbb{E}(\theta, \mathcal{W})$ in linearithmic time. The
alternative search strategy outlined in Algorithm~\ref{alg:studentcnn_leap} can
be easily employed to solve Eq.~\ref{eqn:spladvise}. Step 11 of
Algorithm~\ref{alg:studentcnn_leap} generalizes the three conditions for
selecting samples in terms of both the easiness and the true diverseness. (1)
During training, ``easy'' samples are selected when $\mathcal{L}_{ce}(y_i^{k},
f(x_i^{k},\theta)) < \lambda$, thus $\mathcal{W}_i = 1$. Here, SPL-ADVisE
effectively becomes SPL because $\gamma = 0$, hence the smallest losses from a
single cluster are assigned weights. (2) Otherwise in Step 12, $\mathcal{W}_i =
0$ if the $\mathcal{L}_{ce}(y_i^{k}, f(x_i^{k},\theta)) > \lambda + \gamma$,
which represents the ``hard'' samples with higher losses. (3) We select other
samples by ranking a sample with respect to its loss value within its cluster,
denoted by $i$. Then, we compare the losses to a threshold $\lambda +
\gamma\frac{1}{\sqrt{i}+\sqrt{i-1}}$. Step 11 penalizes samples repeatedly
selected from the same cluster, seeing as this threshold decreases as the
sample's rank $i$ grows. The $\lambda$ and $\gamma$ pace parameters are updated
at the end of training iteration (Step 16). These two pacing parameters are
empirically found by the statistics collected from ranked samples, which is more
robust than by setting absolute values for SPL-ADVisE. This is because setting
by absolute values can result in selecting too many or too few samples.
Specifically, $\lambda$ and $\gamma$ are set according to the ranked samples,
and then absolute values are calculated accordingly. The two pace adaptation
parameters $\{\beta_1,\beta_2\}$ in Algorithm~\ref{alg:studentcnn_leap} are
factors that are used to introduce increasingly hard and diverse samples from
one iteration to the next.

Compared to the original loss function $\mathcal{L}_{ce}$, it is evident that
there is a suppressing effect of the latent SPL loss function on large
losses~\citep{MENG2017319}. In SPL-ADVisE, we embed the true diverseness prior,
in addition to easiness. We can encode these two priors into the latent
variables as a regularizer and avoid unreasonable local minima. This allows us
to select high-confidence samples that are dispersed across the input space,
incrementally learning meaningful knowledge through robust guidance. As a
result, we produce a training scheme that exploits a diverse curriculum
consisting of easy samples from multiple clusters within a learned
representation space.

\section{Experiments}\label{sec:experiments}
All experiments were conducted using the PyTorch framework, while leveraging
containerized multi-GPU training on NVIDIA P100 Pascal GPUs through Docker. We
compared our SPL-ADVisE framework against the original SPLD algorithm and Random
Sampling on FashionMNIST, SVHN, CIFAR-10 and CIFAR-100. We tried 5 different
combinations of $\{\beta_1, \beta_2\}$ for SPL-ADVisE, and determined that
$\beta_1 = 0.1$ and $\beta_2 = 0.1$ provided the best combination with the
smallest error rate. The embedding CNN is trained in parallel with the student
CNN. The computational requirement of training the embedding CNN is mitigated by
leveraging multiprocessing for parallel computing to share data between
processes locally using arrays and values. In our experiments, we mainly compare
convergence in terms of number of mini-batches required to achieve a comparable
state-of-the-art test performance. We also visualize the original
high-dimensional representations using t-SNE~\citep{maaten2008visualizing},
where the different colours correspond to different classes and the values to
density estimates. \\

\noindent \textbf{FashionMNIST}. In the experiments for
FashionMNIST~\citep{xiao2017/online}, we extract feature embeddings from a
LeNet~\citep{lecun-gradientbased-learning-applied-1998} as the embedding CNN,
and learn a representation space using the Magnet Loss. The fully-connected
layer of the LeNet is replaced with an embedding layer for compacting the
distribution of the learned features for feature similarity comparison using the
Magnet Loss. The student CNN (classifier) was a
ResNet-18~\citep{DBLP:conf/cvpr/HeZRS16}, which we then trained with our
SPL-ADVisE framework. The embedding CNN was trained with mini-batches of size 64
and optimized using Adam with a learning rate of 0.0001. The classifier is
trained using SGD with Nesterov momentum of 0.9, weight decay of 0.0005, and a
learning rate of 0.001. We performed data augmentation on the FashionMNIST
training set with normalization, random horizontal flip, random vertical flip,
random translation, random crop, and random rotation. As shown in
Table~\ref{tab:results}, the SPL-ADVisE framework results in an increase in test
accuracy by between 0.95 and 1.35 percentage points on FashionMNIST.\\

\begin{wrapfigure}{R}{0.40\textwidth}
\centering
\subfigure[]{\includegraphics[width=0.32\linewidth]{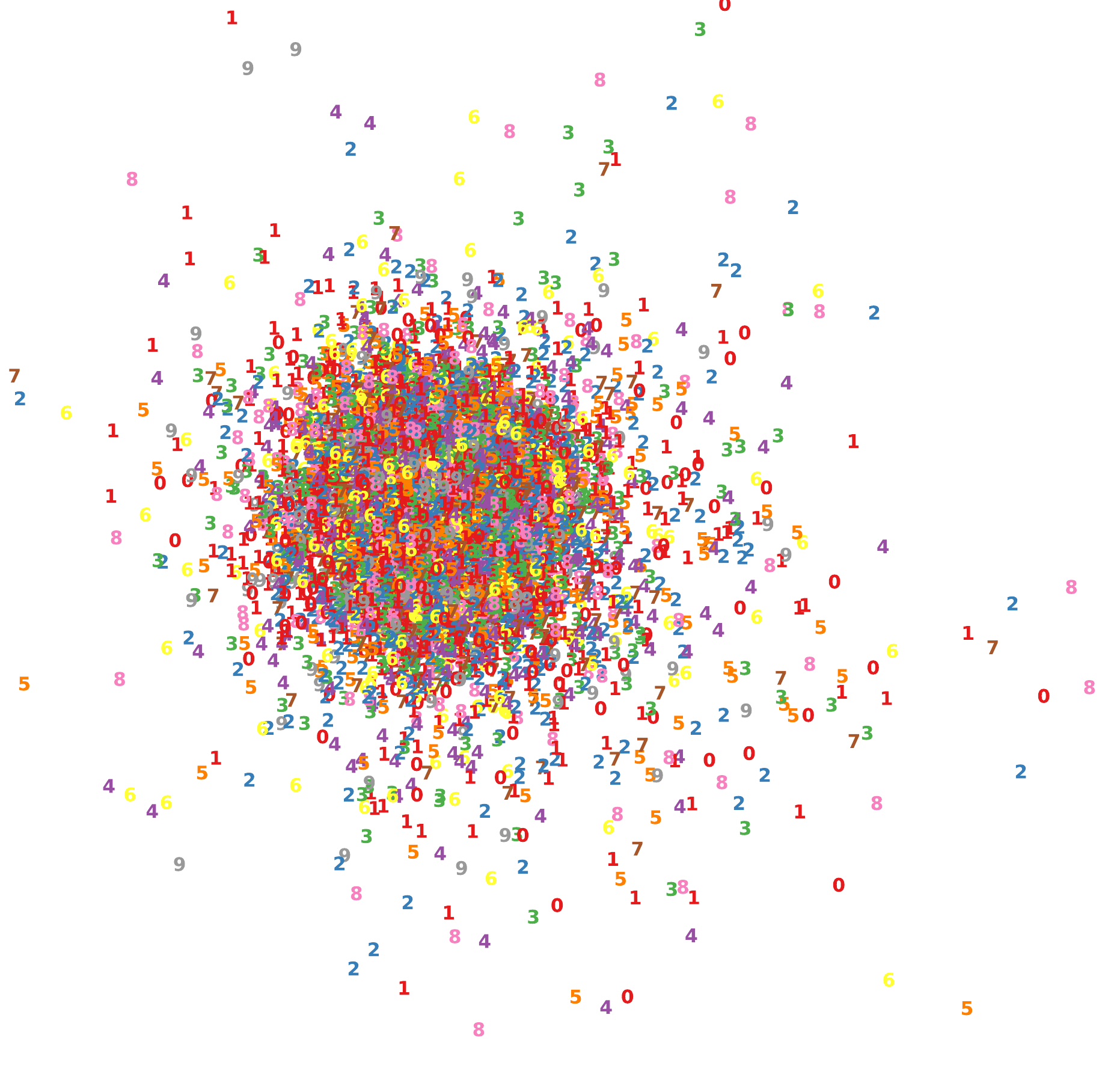}}
\subfigure[]{\includegraphics[width=0.32\linewidth]{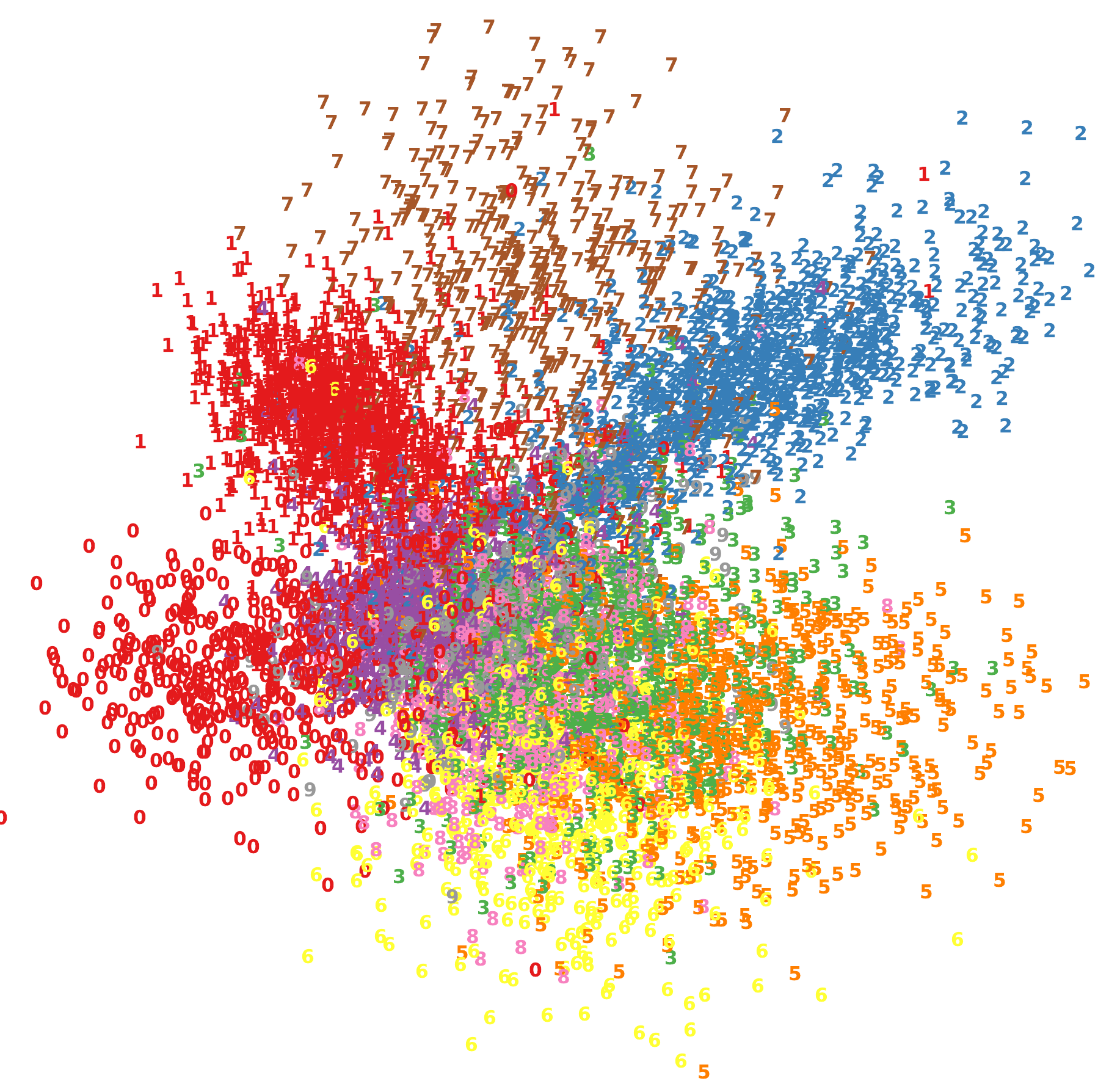}}
\subfigure[]{\includegraphics[width=0.32\linewidth]{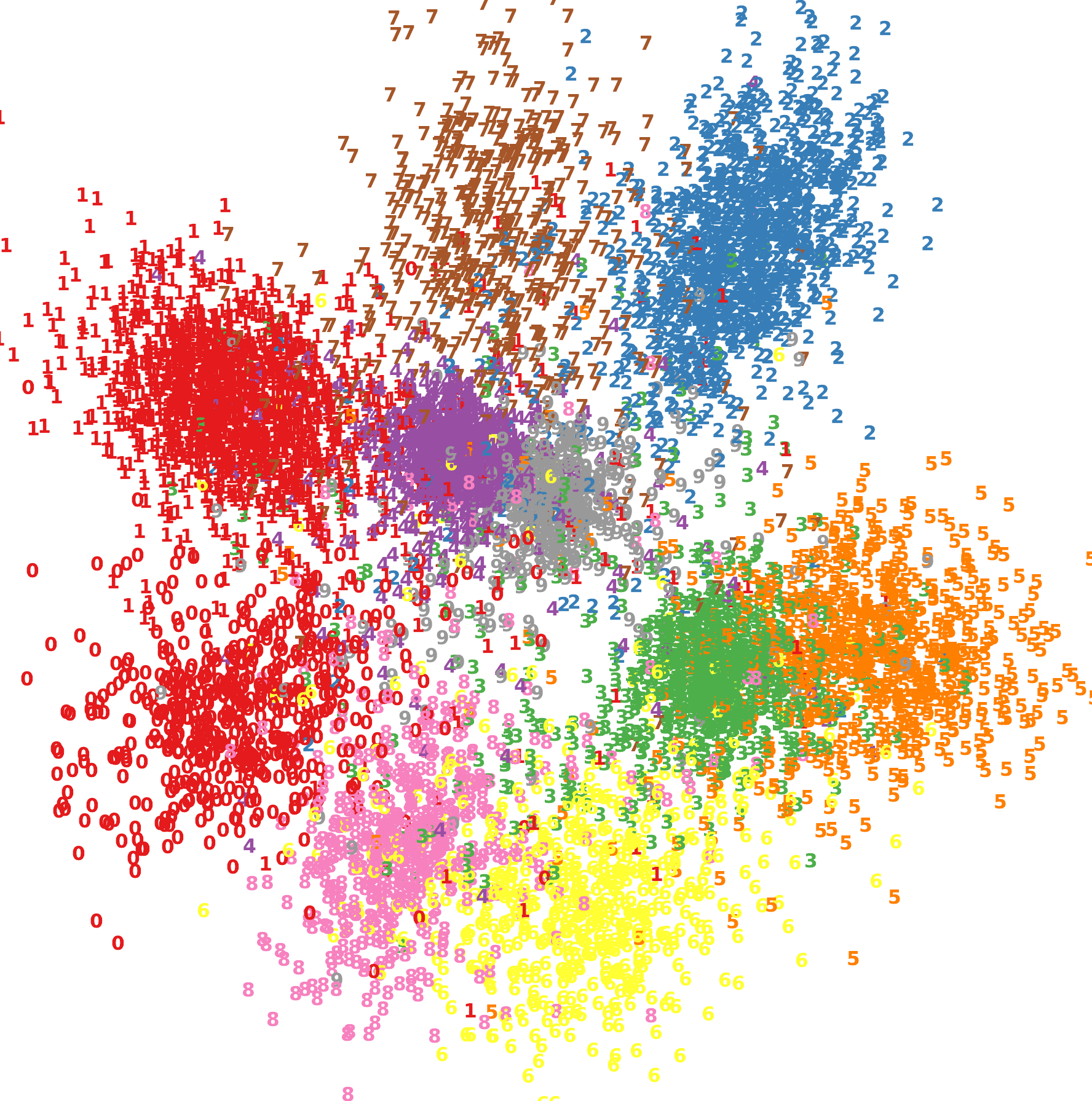}\label{fig:svhnrep}}
\caption{t-SNE visualizations of the SVHN representation space at different stages of training the embedding CNN. (a) Start of training. (b) 6000 mini-batch updates. (c) Final representation space.}
\end{wrapfigure}

\begin{figure}[ht]
\captionsetup{font=small}
\centering
\subfigure[]{\includegraphics[width=0.31\linewidth]{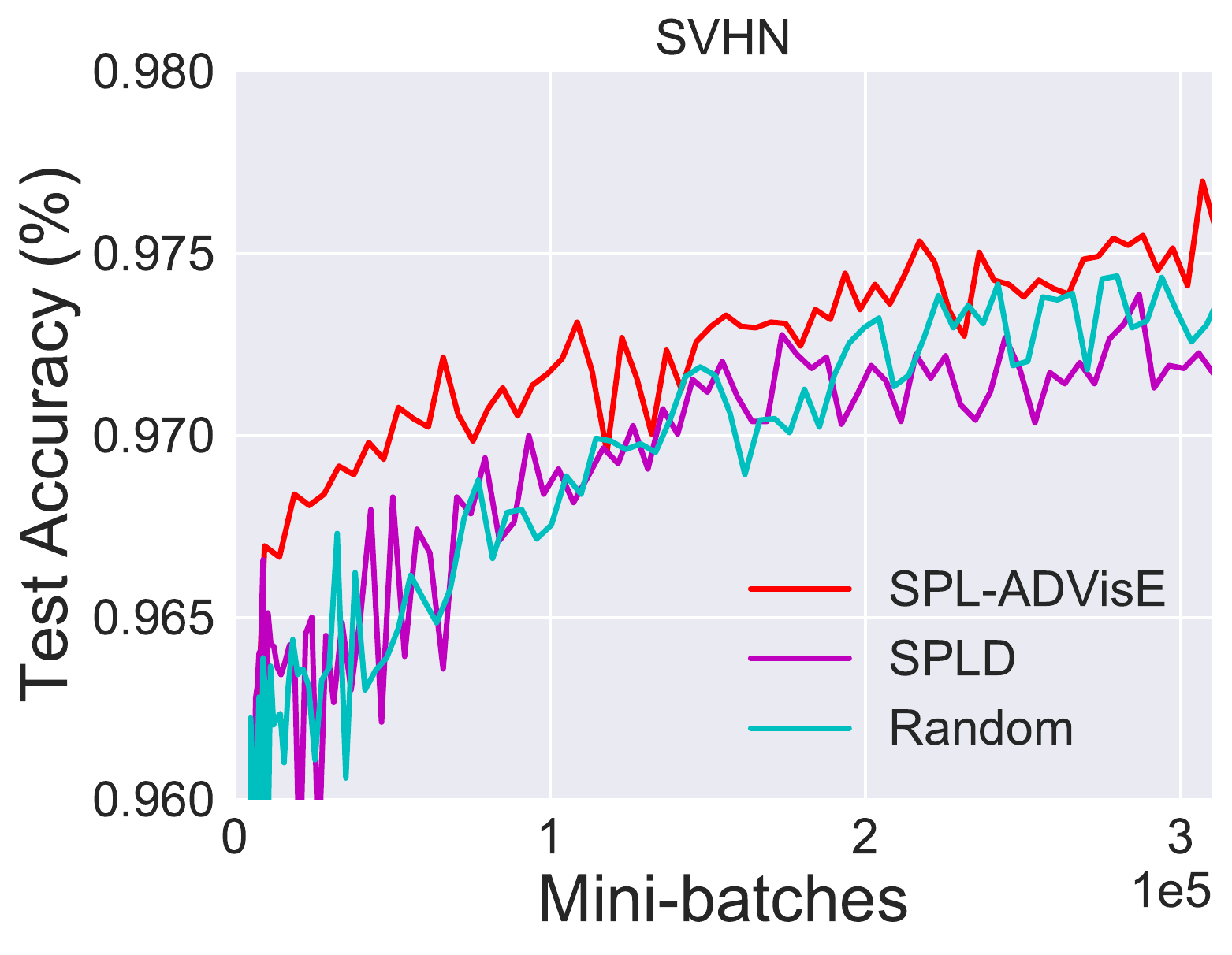}\label{fig:svhntest}}
\subfigure[]{\includegraphics[width=0.31\linewidth]{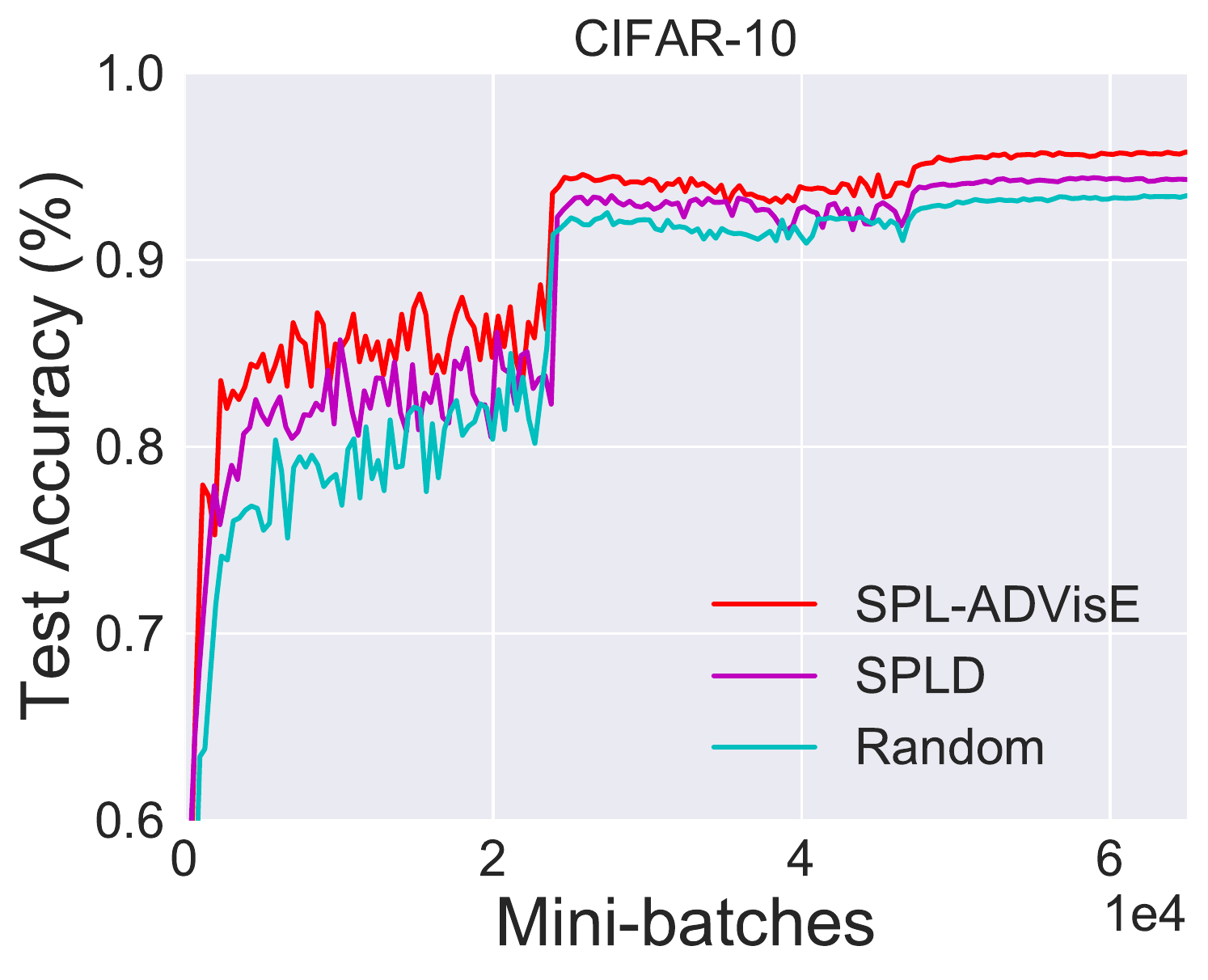}\label{fig:cifar10test}}
\subfigure[]{\includegraphics[width=0.31\linewidth]{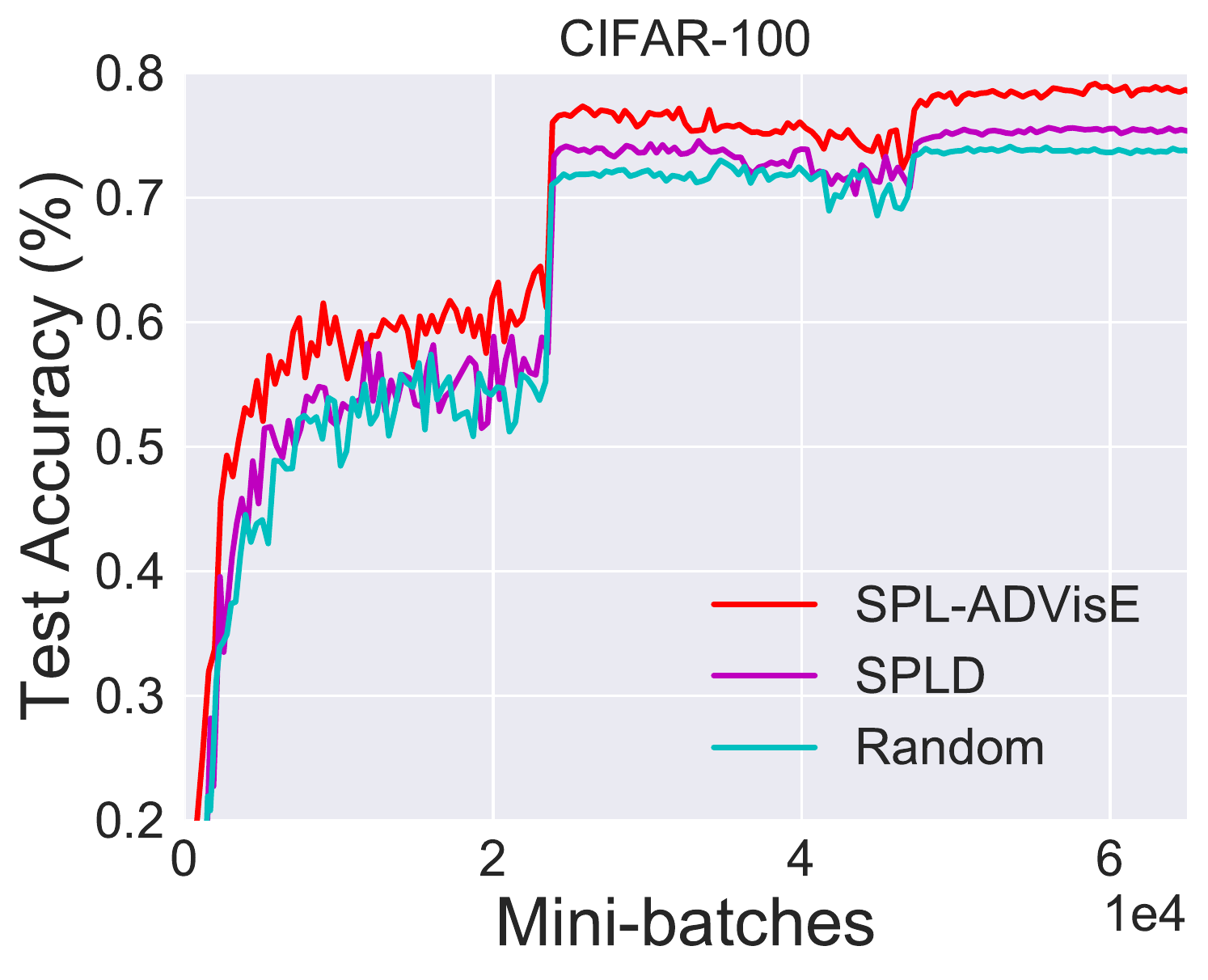}\label{fig:cifar100test}}
\caption{Averaged test curves of 5 independent runs on (a) SVHN, (b) CIFAR-10 with standard data augmentation, and (c) CIFAR-100. The solid line represents our SPL-ADVisE framework.}
\label{fig:alltestcurves}
\end{figure}

\noindent \textbf{SVHN}. The Street View House Numbers (SVHN)
dataset~\citep{svhn} is a real-world image dataset with 630,420 RGB images of
32$\times$32 pixels in size, where each image consists of digits that are from
one of ten different classes. The SVHN dataset is split into the training set,
testing set, and extra set with 73,257, 26,032, and 531,131 images, respectively.
The student CNN used to train on the SVHN dataset was a WideResNet with a fixed
depth of 16, a fixed widening factor of 8, and dropout probability of 0.4. The
SVHN training set and extra set were combined for a total of 604,388 images to
train the student CNN for 65 epochs and no data augmentation scheme was applied.
The student model was optimized using SGD with Nesterov momentum of 0.9, weight
decay of 0.0005, and batch-size of 128.  Our experiments revealed that
VGG-16~\citep{DBLP:journals/corr/SimonyanZ14a} learned strong and rich feature
representations, which yielded the best convergence on the Magnet Loss.
Therefore, we treat the VGG-16 model as a feature extraction engine and use it
for feature extraction from SVHN images, without any fine-tuning. The VGG-16
network is trained with randomly sampled mini-batches of size 64 and optimized
using Adam at a learning rate of 0.0001. The learned feature representation space of
the SVHN data is depicted in Figure \ref{fig:svhnrep}. The results in Figure
\ref{fig:svhntest} show that SPL-ADVisE is able to converge faster to a higher
test accuracy than Random Sampling and SPLD.\\

\noindent \textbf{CIFAR-10}. The CIFAR-10
dataset~\citep{Krizhevsky2009LearningML} consists of 50,000 examples in the
training set and 10,000 examples in the test set. Each example is a 32$\times$32
RGB image associated with a label from 10 classes. The CIFAR-10 training set was
augmented with normalization, random horizontal flip, and random crop. We used
VGG-16 as our embedding CNN and ResNet-18 as our classifier in the CIFAR-10
experiments. We chose ResNet-18 over other architectures because it is faster to
train and achieves good performance on CIFAR-10. The embedding CNN was a VGG-16
setup similar to our SVHN experiments. The CIFAR-10 experiment followed the
learning rate scheduler identical to that of the
WideResNet~\citep{Zagoruyko2016WRN} training scheme. The classifier is trained
with batch sizes of 128, using SGD with a momentum of 0.9, weight decay of
0.0005, and a starting learning rate of 0.1 which is dropped by a factor of 0.1
at 60, 120 and 160 epochs. In our experiments, this would translate to 23,400,
46,800, and 62,400 mini-batch updates. Our CIFAR-10  experiments with the
learning rate scheduler shows that SPL-ADVisE converges faster earlier on in
training compared to SPLD and Random Sampling (Figure~\ref{fig:cifar10test}). As
the learning rate drops by a factor of 0.1 at 23,400, 46,800, and 62,400
mini-batch updates, the classifier under the SPL-ADVisE training protocol
improves test performance by 1.17 and 2.22 percentage points over SPLD and
Random Sampling, respectively. \\

\noindent \textbf{CIFAR-100}. The CIFAR-100
dataset~\citep{Krizhevsky2009LearningML} contains 100 classes and the same
number of samples in the training/test set as CIFAR-10. We evaluated our
SPL-ADVisE framework on CIFAR-100 with a WideResNet (student CNN) that has a
fixed depth of 28, a fixed widening factor of 10, and dropout probability of
0.3. The embedding CNN was a VGG-16 setup similar to our SVHN and CIFAR-10
experiments. The optimizer and learning rate scheduler used for training the
WideResNet was identical to the CIFAR-10 experiments. A data augmentation scheme
was not applied on the CIFAR-100 dataset. The results in
Figure~\ref{fig:cifar100test} reveal that on CIFAR-100, there are noticeable
gains in test performance over baseline methods. SPL-ADVisE was able to increase
test accuracy by between 3.81 and 4.75 percentage points.

\begin{table*}[ht]
\captionsetup{font=small}
\centering
\resizebox{0.87\textwidth}{!}{%
\begin{tabular}{ccccc}
\hline
\hline
Method              &  FashionMNIST  & SVHN*          &
CIFAR-10       & CIFAR-100*     \\ \hline
\textbf{SPL-ADVisE} &  \textbf{94.12 $\pm$ 0.11} &
\textbf{97.68 $\pm$ 0.09} & \textbf{95.48 $\pm$ 0.16} & \textbf{79.17 $\pm$
0.24} \\
SPLD                & 93.17 $\pm$ 0.18          &
97.38 $\pm$ 0.06          & 94.31 $\pm$ 0.21         & 75.36  $\pm$ 0.30
\\
Random Sampling             & 92.77  $\pm$ 0.14        &
97.41   $\pm$ 0.11       & 93.26  $\pm$ 0.12        & 74.42 $\pm$ 0.25 \\ \hline
\end{tabular}}
\caption{Experimental results across all datasets (FashionMNIST,
CIFAR-10, CIFAR-100, and SVHN) and sampling methods (SPL-ADVisE, SPLD, and
Random Sampling). The test accuracy (\%) results are averaged over 5 runs and ``*''
indicates that no data augmentation scheme was applied on the dataset.}
\label{tab:results}
\end{table*}

\section{Discussion}
We have shown that fusing a salient non-linear representation space with a
dynamic learning strategy can help a DNN converge towards an optimal solution. A
random curriculum or a dynamic learning strategy without a good representation
space was found to achieve a lower test accuracy or converge more slowly than
SPL-ADVisE. Biasing samples based on the \textit{easiness} and \textit{true
diverseness} to select mini-batches shows improvement in convergence to achieve
classification performance comparable or better than the baselines, Random Sampling and
SPLD. As shown in Table~\ref{tab:results}, the student CNN models show increased
accuracy on FashionMNIST, SVHN, CIFAR-10, and CIFAR-100 with our SPL-ADVisE
sampling method. It is to be noted that the SPL-ADVisE framework improves the
performance of complex convolutional architectures which already leverage
regularization techniques such as batch normalization, dropout, and data
augmentation. We see that the improvements on coarse-grained datasets such as
FashionMNIST, SVHN, and CIFAR-10 are between 0.27 and 2.22 percentage points.
On a fine-grained dataset like CIFAR-100, it is more challenging to obtain a
high classification accuracy. This is because there are a 100 fine-grained
classes but the number of training instances for each class is small. We have
only 500 training images and 100 testing images per class. In addition, the
dataset contains images of low quality and images where only part of the object
is visible (i.e.~for a person, only head or only body). However, we show that
with SPL-ADVisE, we can attain a significant increase in accuracy by 3.81 and
4.75 percentage points over the baselines SPLD and Random Sampling, respectively. The mix
of easy and diverse samples from a more accurate representation space of the
data helps select appropriate samples during different stages of training and
guide the network to achieve a higher classification accuracy, especially for
more difficult fine-grained classification tasks.

The embedding CNN that leverages the Magnet Loss is used to obtain the \textit{true diverseness}
because, unlike other DML loss functions, the Magnet Loss forms clusters. This
made it a natural choice, seeing as we need clusters for our notion of diversity
derived from SPLD. The Magnet Loss is designed to operate on entire regions of
the embedding space that the examples inhabit. Moreover, the Magnet Loss learns
to model the distributions of different classes in the representation space and
reduces local distribution overlap. Although we propose to use the Magnet Loss
to train the embedding CNN, it is still possible to use other DML loss
functions, as long as they are capable of forming clusters in the representation
space. To summarize, we do not restrict the embedding network to utilize a
particular deep network objective or architecture. It is designed to be modular
such that we can experiment with different DML techniques that could provide
optimal clustering results. Also, the student network does not depend on what
type of embedding network we use, as long as it can determine the diversity
prior based on clusters identified in the representation space.

\section{Conclusion}
We introduced SPL-ADVisE, an end-to-end representation learning SPL strategy
for adaptive
mini-batch formation. Our method uses an embedding CNN for learning an
expressive representation space through a DML technique called the Magnet Loss.
The student CNN is a classifier that can exploit this new knowledge from the
representation space to place the \textit{true diverseness} and
\textit{easiness} as sample importance priors during online mini-batch
selection. The computational overhead of training two CNNs can be mitigated by
training the embedding CNN and student CNN in parallel. SPL-ADVisE achieves
good convergence speed and higher test
performance on FashionMNIST, SVHN, CIFAR-10, and CIFAR-100 using a
combination of two CNN
architectures. We hope this will help foster progress of end-to-end SPL fused
DML strategies for DNN training, where a number of potentially interesting
directions can be considered for further exploration. Our framework is
implemented in PyTorch and will be released as open-source on GitHub at \url{https://github.com/vithursant/SPL-ADVisE}. \\

\noindent \textbf{Acknowledgement}
The authors acknowledge financial support by the Natural Sciences and
Engineering Research Council of Canada (NSERC), the Canada Foundation for
Innovation (CFI), and an Amazon Academic Research Award. The authors wish to
acknowledge the hardware support from NVIDIA for donated GPUs. We thank Colin Brennan
for helpful edits and revision recommendations that improved the overall
presentation of our manuscript. \clearpage

\bibliography{refs}

\begin{thebibliography}{33}
\providecommand{\natexlab}[1]{#1}
\providecommand{\url}[1]{\texttt{#1}}
\expandafter\ifx\csname urlstyle\endcsname\relax
  \providecommand{\doi}[1]{doi: #1}\else
  \providecommand{\doi}{doi: \begingroup \urlstyle{rm}\Url}\fi

\bibitem[Arthur and Vassilvitskii(2007)]{Arthur:2007:KAC:1283383.1283494}
David Arthur and Sergei Vassilvitskii.
\newblock K-means++: The advantages of careful seeding.
\newblock In \emph{Proceedings of the Eighteenth Annual ACM-SIAM Symposium on
  Discrete Algorithms}, SODA '07, pages 1027--1035, 2007.

\bibitem[Bell and Bala(2015)]{bell15productnet}
Sean Bell and Kavita Bala.
\newblock Learning visual similarity for product design with convolutional
  neural networks.
\newblock \emph{ACM Trans. Graph.}, 34\penalty0 (4):\penalty0 98:1--98:10, July
  2015.

\bibitem[Bengio et~al.(2009)Bengio, Louradour, Collobert, and
  Weston]{Bengio:2009:CL:1553374.1553380}
Yoshua Bengio, J{\'e}r\^{o}me Louradour, Ronan Collobert, and Jason Weston.
\newblock Curriculum learning.
\newblock In \emph{International Conference on Machine Learning}, pages 41--48,
  2009.

\bibitem[Choy et~al.(2016)Choy, Gwak, Savarese, and Chandraker]{choy_nips16}
Christopher~Bongsoo Choy, JunYoung Gwak, Silvio Savarese, and Manmohan~Krishna
  Chandraker.
\newblock Universal correspondence network.
\newblock In \emph{Advances in Neural Information Processing Systems}, pages
  2406--2414, 2016.

\bibitem[Elman(1993)]{Elman1993-ELMLAD}
Jeffrey~L. Elman.
\newblock Learning and development in neural networks: the importance of
  starting small.
\newblock \emph{Cognition}, 48\penalty0 (1):\penalty0 71 -- 99, 1993.

\bibitem[Frome et~al.(2013)Frome, Corrado, Shlens, Bengio, Dean, Ranzato, and
  Mikolov]{41869}
Andrea Frome, Gregory~S. Corrado, Jonathon Shlens, Samy Bengio, Jeffrey Dean,
  Marc'Aurelio Ranzato, and Tomas Mikolov.
\newblock Devise: A deep visual-semantic embedding model.
\newblock In \emph{Advances in Neural Information Processing Systems}, pages
  2121--2129, 2013.

\bibitem[G{\"{u}}l{\c{c}}ehre and Bengio(2016)]{Glehre2016KnowledgeMI}
{\c{C}}aglar G{\"{u}}l{\c{c}}ehre and Yoshua Bengio.
\newblock Knowledge matters: Importance of prior information for optimization.
\newblock \emph{Journal of Machine Learning Research}, 17:\penalty0 8:1--8:32,
  2016.

\bibitem[He et~al.(2016)He, Zhang, Ren, and Sun]{DBLP:conf/cvpr/HeZRS16}
Kaiming He, Xiangyu Zhang, Shaoqing Ren, and Jian Sun.
\newblock Deep residual learning for image recognition.
\newblock In \emph{Proceedings of the IEEE Conference on Computer Vision and
  Pattern Recognition}, pages 770--778, 2016.

\bibitem[Huang et~al.(2017)Huang, Gu, Ma, and Li]{huang2017self}
Wenhui Huang, Jason Gu, Xin Ma, and Yibin Li.
\newblock Self-paced model learning for robust visual tracking.
\newblock \emph{Journal of Electronic Imaging}, 26\penalty0 (1):\penalty0
  13016, 2017.

\bibitem[Im and Taylor(2016)]{Im2016ccml}
Daniel~Jiwoong Im and Graham~W. Taylor.
\newblock Learning a metric for class-conditional {KNN}.
\newblock In \emph{International Joint Conference on Neural Networks}, pages
  1932--1939, 2016.

\bibitem[Jiang et~al.(2014)Jiang, Meng, Yu, Lan, Shan, and
  Hauptmann]{Jiang2014SelfPacedLW}
Lu~Jiang, Deyu Meng, Shoou{-}I Yu, Zhen{-}Zhong Lan, Shiguang Shan, and
  Alexander~G. Hauptmann.
\newblock Self-paced learning with diversity.
\newblock In \emph{Advances in Neural Information Processing Systems}, pages
  2078--2086, 2014.

\bibitem[Khan et~al.(2011)Khan, Zhu, and Mutlu]{Khan2011HowDH}
Faisal Khan, Xiaojin~(Jerry) Zhu, and Bilge Mutlu.
\newblock How do humans teach: On curriculum learning and teaching dimension.
\newblock In \emph{Advances in Neural Information Processing Systems}, pages
  1449--1457, 2011.

\bibitem[Krizhevsky and Hinton(2009)]{Krizhevsky2009LearningML}
A.~Krizhevsky and G.~Hinton.
\newblock Learning multiple layers of features from tiny images.
\newblock \emph{Master's thesis, Department of Computer Science, University of
  Toronto}, 2009.

\bibitem[Kumar et~al.(2010)Kumar, Packer, and
  Koller]{Kumar:2010:SLL:2997189.2997322}
M.~Pawan Kumar, Benjamin Packer, and Daphne Koller.
\newblock Self-paced learning for latent variable models.
\newblock In \emph{Advances in Neural Information Processing Systems}, pages
  1189--1197, 2010.

\bibitem[Lapedriza et~al.(2013)Lapedriza, Pirsiavash, Bylinskii, and
  Torralba]{lapedriza2013all}
{\`{A}}gata Lapedriza, Hamed Pirsiavash, Zoya Bylinskii, and Antonio Torralba.
\newblock Are all training examples equally valuable?
\newblock \emph{CoRR}, abs/1311.6510, 2013.
\newblock URL \url{http://arxiv.org/abs/1311.6510}.

\bibitem[LeCun et~al.(2001)LeCun, Bottou, Bengio, and
  Haffner]{lecun-gradientbased-learning-applied-1998}
Yann LeCun, Leon Bottou, Yoshua Bengio, and Patrick Haffner.
\newblock Gradient-based learning applied to document recognition.
\newblock In \emph{{IEEE} Intelligent Signal Processing}, pages 306--351. 2001.

\bibitem[Lee and Grauman(2011)]{Lee2011LearningTE}
Yong~Jae Lee and Kristen Grauman.
\newblock Learning the easy things first: Self-paced visual category discovery.
\newblock In \emph{Proceedings of the {IEEE} Conference on Computer Vision and
  Pattern Recognition}, pages 1721--1728, 2011.

\bibitem[Li et~al.(2017)Li, Zhong, Lin, Guo, Sun, Sitek, Ye, Thrall, and
  Li]{li2017self}
Xiang Li, Aoxiao Zhong, Ming Lin, Ning Guo, Mu~Sun, Arkadiusz Sitek, Jieping
  Ye, James Thrall, and Quanzheng Li.
\newblock Self-paced convolutional neural network for computer aided detection
  in medical imaging analysis.
\newblock In \emph{MLMI@MICCAI}, 2017.

\bibitem[Loshchilov and Hutter(2015)]{DBLP:journals/corr/LoshchilovH15}
Ilya Loshchilov and Frank Hutter.
\newblock Online batch selection for faster training of neural networks.
\newblock \emph{CoRR}, abs/1511.06343, 2015.
\newblock URL \url{http://arxiv.org/abs/1511.06343}.

\bibitem[Meng et~al.(2017)Meng, Zhao, and Jiang]{MENG2017319}
Deyu Meng, Qian Zhao, and Lu~Jiang.
\newblock A theoretical understanding of self-paced learning.
\newblock \emph{Information Sciences}, 414:\penalty0 319 -- 328, 2017.

\bibitem[Netzer et~al.(2011)Netzer, Wang, Coates, Bissacco, Wu, and Ng]{svhn}
Yuval Netzer, Tao Wang, Adam Coates, Alessandro Bissacco, Bo~Wu, and Andrew~Y.
  Ng.
\newblock Reading digits in natural images with unsupervised feature learning.
\newblock In \emph{NIPS Workshop on Deep Learning and Unsupervised Feature
  Learning}, 2011.

\bibitem[Peng(2017)]{DBLP:conf/atal/Peng17}
Bei Peng.
\newblock How do humans teach: On curriculum design for machine learners.
\newblock In \emph{Proceedings of the Conference on Autonomous Agents and
  MultiAgent Systems}, pages 1851--1852, 2017.

\bibitem[Rippel et~al.(2016)Rippel, Paluri, Doll{\'{a}}r, and
  Bourdev]{DBLP:journals/corr/RippelPDB15}
Oren Rippel, Manohar Paluri, Piotr Doll{\'{a}}r, and Lubomir~D. Bourdev.
\newblock Metric learning with adaptive density discrimination.
\newblock In \emph{International Conference on Learning Representations}, 2016.

\bibitem[Schroff et~al.(2015)Schroff, Kalenichenko, and
  Philbin]{Schroff2015FaceNetAU}
Florian Schroff, Dmitry Kalenichenko, and James Philbin.
\newblock Facenet: {A} unified embedding for face recognition and clustering.
\newblock In \emph{Proceedings of the {IEEE} Conference on Computer Vision and
  Pattern Recognition}, pages 815--823, 2015.

\bibitem[Simonyan and Zisserman(2014)]{DBLP:journals/corr/SimonyanZ14a}
Karen Simonyan and Andrew Zisserman.
\newblock Very deep convolutional networks for large-scale image recognition.
\newblock \emph{CoRR}, abs/1409.1556, 2014.

\bibitem[Sohn(2016)]{Sohn2016ImprovedDM}
Kihyuk Sohn.
\newblock Improved deep metric learning with multi-class n-pair loss objective.
\newblock In \emph{Advances in Neural Information Processing Systems}, pages
  1849--1857, 2016.

\bibitem[Song et~al.(2016)Song, Xiang, Jegelka, and Savarese]{songCVPR16}
Hyun~Oh Song, Yu~Xiang, Stefanie Jegelka, and Silvio Savarese.
\newblock Deep metric learning via lifted structured feature embedding.
\newblock In \emph{Proceedings of the {IEEE} Conference on Computer Vision and
  Pattern Recognition}, pages 4004--4012, 2016.

\bibitem[Song et~al.(2017)Song, Jegelka, Rathod, and Murphy]{Song2017DeepML}
Hyun~Oh Song, Stefanie Jegelka, Vivek Rathod, and Kevin Murphy.
\newblock Deep metric learning via facility location.
\newblock In \emph{Proceedings of the {IEEE} Conference on Computer Vision and
  Pattern Recognition}, pages 2206--2214, 2017.

\bibitem[van~der Maaten and Hinton(2008)]{maaten2008visualizing}
Laurens van~der Maaten and Geoffrey~E. Hinton.
\newblock Visualizing high-dimensional data using t-sne.
\newblock \emph{Journal of Machine Learning Research}, 9:\penalty0 2579--2605,
  2008.

\bibitem[Wang et~al.(2017)Wang, Zhou, Wen, Liu, and Lin]{Wang2017DeepML}
Jian Wang, Feng Zhou, Shilei Wen, Xiao Liu, and Yuanqing Lin.
\newblock Deep metric learning with angular loss.
\newblock In \emph{Proceedings of the {IEEE} Conference on Computer Vision},
  pages 2612--2620, 2017.

\bibitem[Xiao et~al.(2017)Xiao, Rasul, and Vollgraf]{xiao2017/online}
Han Xiao, Kashif Rasul, and Roland Vollgraf.
\newblock Fashion-mnist: a novel image dataset for benchmarking machine
  learning algorithms.
\newblock \emph{CoRR}, abs/1708.07747, 2017.
\newblock URL \url{http://arxiv.org/abs/1708.07747}.

\bibitem[Zagoruyko and Komodakis(2016)]{Zagoruyko2016WRN}
Sergey Zagoruyko and Nikos Komodakis.
\newblock Wide residual networks.
\newblock In \emph{Proceedings of the British Machine Vision Conference}, 2016.

\bibitem[Zhou et~al.(2018)Zhou, Wang, Meng, Xin, Li, Gong, and
  Zheng]{Zhou2017DeepSL}
Sanping Zhou, Jinjun Wang, Deyu Meng, Xiaomeng Xin, Yubing Li, Yihong Gong, and
  Nanning Zheng.
\newblock Deep self-paced learning for person re-identification.
\newblock \emph{Journal of Pattern Recognition}, 76:\penalty0 739--751, 2018.

\end{thebibliography}
\end{document}